\documentclass{article}
\usepackage{ijcai23}

\usepackage{times}
\usepackage{soul,color,xcolor}
\usepackage{url}
\usepackage[hidelinks]{hyperref}
\usepackage[utf8]{inputenc}
\usepackage[small]{caption}
\usepackage{graphicx}
\usepackage{amsmath}
\usepackage{amsthm}
\usepackage{amssymb}
\usepackage{booktabs}
\usepackage{multirow}
\usepackage{algorithm}
\usepackage{algorithmic}
\usepackage[switch]{lineno}

\title{DenseDINO: Boosting Dense Self-Supervised Learning with Token-Based Point-Level Consistency}

\author{
Yike Yuan$^1$
\and
Xinghe Fu$^1$\and
Yunlong Yu$^2$\And
Xi Li$^{13}$\thanks{The corresponding author is Xi Li.}
\affiliations
$^1$College of Computer Science and Technology, Zhejiang University\\
$^2$College of Information Science and Electronic Engineering, Zhejiang University\\
$^3$Zhejiang – Singapore Innovation and AI Joint Research Lab, Hangzhou\\
\emails
\{yuanyike, xinghefu, yuyunlong, xilizju\}@zju.edu.cn
}

\begin{document}

\maketitle

\begin{abstract}
In this paper, we propose a simple yet effective transformer framework for self-supervised learning called DenseDINO to learn dense visual representations. To exploit the spatial information that the dense prediction tasks require but neglected by the existing self-supervised transformers, we introduce point-level supervision across views in a novel token-based way. Specifically, DenseDINO introduces some extra input tokens called reference tokens to match the point-level features with the position prior. With the reference token, the model could maintain spatial consistency and deal with multi-object complex scene images, thus generalizing better on dense prediction tasks. Compared with the vanilla DINO, our approach obtains competitive performance when evaluated on classification in ImageNet and achieves a large margin (+7.2\% mIoU) improvement in semantic segmentation on PascalVOC under the linear probing protocol for segmentation.
\end{abstract}

\section{Introduction}
In recent years, self-supervised learning approaches have achieved great success, narrowed the performance gap with, and even outperformed their supervised counterparts on downstream tasks. Among them, the self-distillation framework DINO \cite{DINO} surpasses all previous methods on different backbones, including a number of variants of CNN and transformer.

DINO adopts a classic pipeline in self-supervised learning: generate multiple augmented views of one image, extract features of each view and maximize their similarity. DINO matches the global view feature, {\it i.e.}, the feature vector obtained by either pooling the feature map of CNN or the output class token of transformer.
Many works\cite{densecl,vader} have pointed out that matching global representations might be sufficient for image-level prediction tasks like classification but not enough for dense prediction tasks like segmentation since the pixel-wise prediction is needed in segmentation but spatial information is discarded when extracting global representation.
Therefore, many approaches were proposed to introduce pixel-level or object-level pretext tasks for learning dense representation in a self-supervised manner.

\begin{figure}[t]
    \centering
    \includegraphics[width=\linewidth]{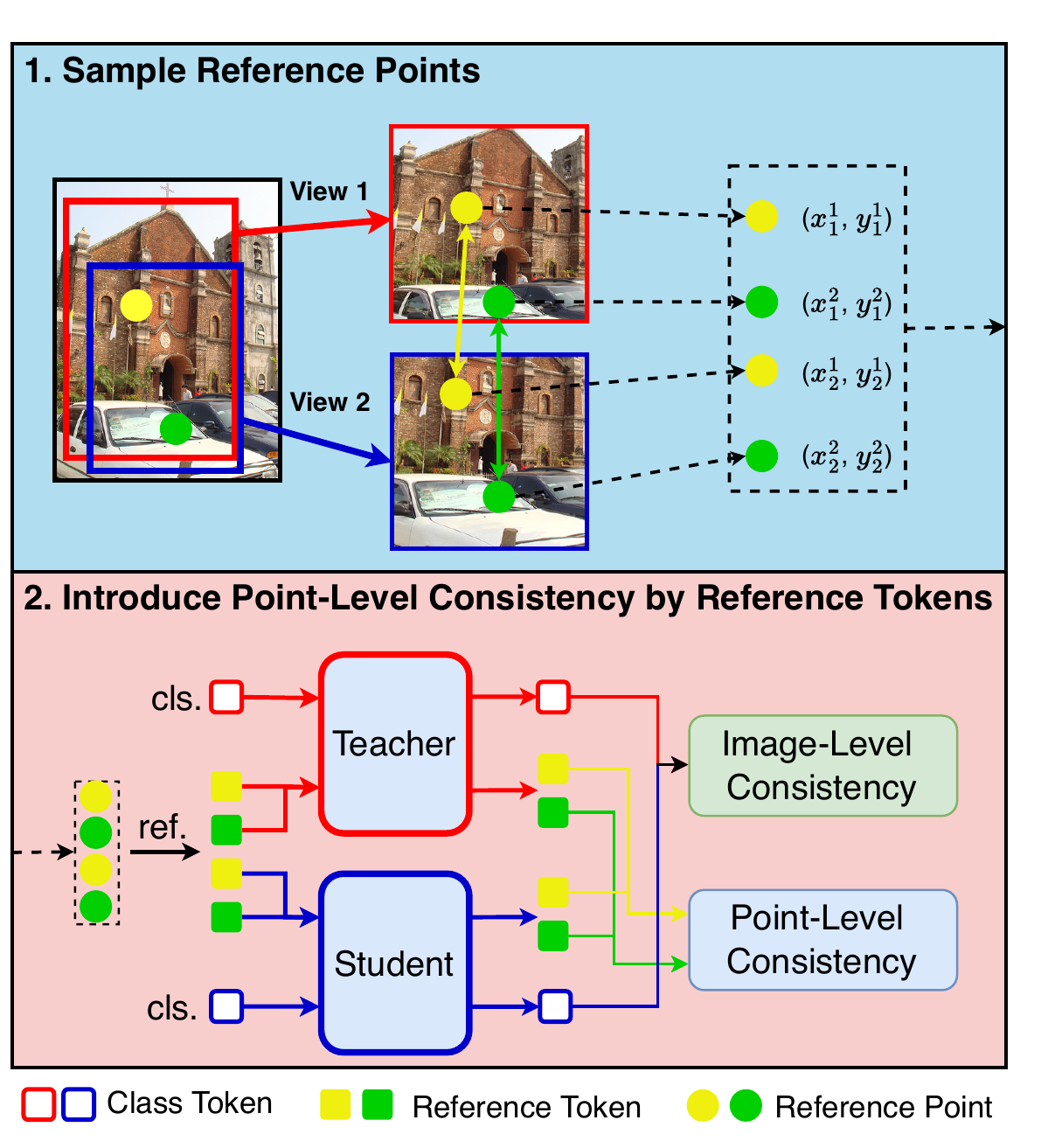}
    \caption{DenseDINO maximizes both image-level and point-level consistency. Multiple reference point pairs are sampled in the pair of image views and their coordinates are then encoded into tokens, named reference tokens, which served as transformer input. We regard the class token as the image-level feature and the reference tokens as the point-level features.}
    \label{fig:top}
\end{figure}

In this paper, we focus on the self-supervised learning on the transformer as its superiority to CNNs on various vision tasks and remedy the existing approaches from the following issues. Firstly, existing approaches treat the output patch tokens as the feature map of the transformer and supervise it directly. We think it is sub-optimal since the ability of the attention mechanism to spread semantic information between tokens is not fully explored. Secondly, the existing self-supervised transformers hardly produce high-quality vision representations that generalize well on different downstream tasks. For example, the competitors Leopart \cite{leopart} and DINO \cite{DINO} hardly perform well simultaneously on both classification and segmentation tasks (Detailed results are in Section \ref{sec:exp}). 

To address the issues mentioned above, on the basis of DINO, we develop a simple yet effective framework, named DenseDINO, which introduces point-level supervision in a token-based way for self-supervised transformers to exploit spatial information. 
As illustrated in Figure \ref{fig:top}, during training, multiple point pairs are randomly sampled across views for point-level supervision. Similar to the class token which aggregates global semantic information, we additionally introduce a series of tokens, named reference tokens, to obtain point-level features. To establish the correspondence between reference tokens and reference points, we define the reference token as the positional embedding of the reference point, computed by interpolating patch positional embedding according to the coordinate of the reference point.

The reference tokens have the following merits. On the one hand, similar to the class token, the reference token could interact with other tokens in the transformer through attention to flexibly gather information. On the other hand, different from the class token that mainly exploits the global semantic information, the reference token plays the role of object query with position prior to attending to local objects and ensures spatial consistency which is essential to dense prediction tasks. Equipped with both class tokens and reference tokens, the transformers could exploit supervision of different granularities, and thus could generalize well on both image-level and dense prediction downstream tasks. 

Besides, it is experimentally observed that the model trained with multi-crop, which adds extra local views that cover small parts of the input image during training, performs better than its counterparts when evaluated on the image-level prediction task like classification but worse on the dense prediction task like segmentation.

Therefore, in this paper, we analyze what multi-crop brings to self-supervised learning. 
We argue that the introduction of local views aggravates the object misalignment between views. On the one hand,  this misalignment helps the model to focus on the salient category-related object and thus benefits classification. 
On the other hand, it also violates spatial consistency and fails to encode multiple objects in a complex scene, resulting in performance degradation on dense prediction tasks. 
Finally, we address the issue by enlarging the scale and resolution of local views and setting a threshold for the overlapped area between views in our DenseDINO to reduce the occurrence of misalignment.

Our main contributions are summarized as follows:
\begin{itemize}
    \item We propose a simple yet effective framework, named DenseDINO, which introduces point-level consistency into DINO in a friendly token-based way for learning dense visual representations.
    \item We explain why multi-crop drops the performances of the dense prediction tasks by analyzing the object misalignment between views and alleviate the issue by modifying the generation mechanism of views.
    \item Our method performs well on both image-level prediction and dense prediction downstream tasks. Compared to the strong baseline DINO\cite{DINO}, our method is competitive when evaluated on the ImageNet classification task and surpasses DINO with a large margin on the PascalVOC segmentation task.
\end{itemize}

\begin{figure*}[ht]
    \centering
    \includegraphics[height=50mm,width=160mm]{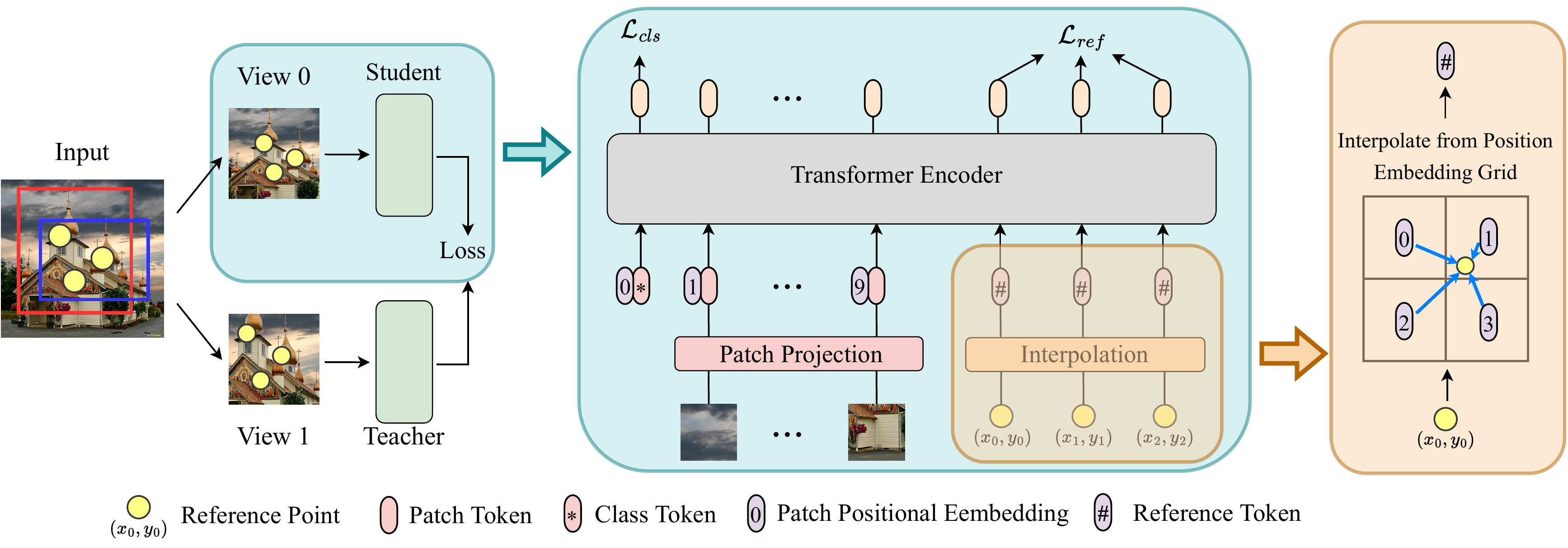}
    \caption{ An overview of our proposed DenseDINO framework. For simplicity, only the case of one pair of views is illustrated in the figure. The input image is augmented into two views and several reference point pairs are sampled in the overlapped area. In preprocessing, the relative coordinate of each point $(x_i, y_i)$ is encoded into a token form by interpolation. Specifically, we compute the reference token, {\it i.e.} the positional embedding of each reference point, by interpolation from the nearby patch positional embeddings. After encoding, we maximize the similarity of the output class token and reference tokens separately.}
    \label{fig:pipeline}
\end{figure*}

\section{Related Work}

\subsection{Image-level Self-supervised Learning}

Self-supervised learning has been widely used as a pre-training approach for obtaining transferable vision representation without human annotations. 
Recently, methods based on contrastive learning\cite{Moco,simclr} have achieved superior performance and even outperformed their supervised counterparts. 
They replace the standard classification with instance discrimination, which discriminates the augmented image in a large image set via a noise contrastive estimator\cite{knn}. 
The quality of the learned feature highly relies on the number of negative samples involved for contrast and thus large memory bank\cite{Moco} or batch of large size\cite{simclr} is needed. 
Methods based on online clustering\cite{sela,swav} address the memory issue by assigning image features to a set of prototypes and comparing the cluster assignments rather than contrasting image features explicitly.
Further, some researchers reformulate the framework in a distillation manner\cite{byol}, which disentangles the training objective from large image sets and finds new ways to avoid collapse without contrasting like stop gradients\cite{simsiam} or centering and sharpening\cite{DINO}.
In fact, all methods mentioned above follow a common framework of three parts: an aggressive data augmentation scheme to generate different views of an image, a siamese network of any backbone for encoding view features, and an optimization objective that matches image features between views and network branches.
Our work follows such a framework on the basis of DINO and leverages the pair views generated from one image to introduce point-level consistency for dense visual representation.

\subsection{Dense Self-supervised Learning}

There remains an issue for image-level self-supervised approaches mentioned in the previous subsection that their potential for dense prediction downstream tasks is not fully exploited.
Therefore, a series of sub-image-level approaches are developed on the basis of their image-level counterparts to maintain the spatial relationship by obtaining fine-grained features during training.
Pixel-level approaches focus on matching each feature vector on the feature map between views. Since image views were cropped, distance metrics including euclidean distance\cite{vader,pixpro} in image space, earth mover's distance\cite{selfemd} and cosine distance\cite{densecl} in feature space, are used to define the correspondence between feature vectors. 
Object-level approaches\cite{maskcontrast,detcon,visconcept} adopt a predetermined mid-level prior, more concretely, object masks predicted by an unsupervised estimator, and contrast object features by masked pooling to avoid misalignment.
And region-level approach\cite{pointcontrast} takes inspiration from object-level approaches to contrast point pairs from different regions and thus improves the robustness to imperfect region assignment.
Combining different levels of supervision in an appropriate way\cite{densesiam,lewel} could help the model generalize better on both image-level prediction and dense prediction downstream tasks.
Moreover, it is worth mentioning that a patch-level approach named Leopart\cite{leopart} is proposed for training self-supervised transformers. Leopart regards each patch as an object and sets a dense clustering task for patch tokens. It achieves state-of-the-art performance on unsupervised segmentation, which suggests the huge potential of the self-supervised transformer for dense prediction tasks.
In this paper, inspired by sub-image-level approaches developed on CNN, we propose a novel token-based strategy that maximizes point-level consistency by patch-level supervision and serves as a complement to those image-level approaches to improve the generalization ability of the learned representation.

\section{Method}

In this section, We firstly review the selected baseline DINO\cite{DINO} for its effectiveness and good compatibility with transformers in Section \ref{sec:bg}, and then we detail the proposed DenseDINO to remedy the baseline from introducing point-level consistency in Section \ref{sec:ref}. Finally, we analyze the influence of multi-crop and give our suggestions in Section \ref{sec:multi-crop}.

\subsection{Review the Baseline DINO}
\label{sec:bg}

DINO\cite{DINO} adopts a siamese network to learn invariant features by matching positive samples in a distillation manner without the requirement of negative samples.
Specifically, DINO includes a student network and a teacher network which have the same architecture. The student is updated online with gradient descent while the teacher is built offline from the student by exponential moving average. Two ingenious operations centering and sharpening are proposed in DINO to avoid feature collapse. 

In the training stage, given an input image $x$, it will be separated and projected into a series of $N$ patch tokens. After encoding by transformer blocks, we got a vector of output tokens $[g_{\theta_s}(x)_{(cls)}, g_{\theta_s}(x)_{(p_0)}, \cdots, g_{\theta_s}(x)_{(p_{N-1})}]$, where $g_{\theta_s}$ denotes the student network, $cls$ denotes the class token in ViT and $p_i$ denotes the $i$-th patch token. The output probability $P$ of the student can be obtained by normalizing the class token with softmax function of temperature $\tau_s$, formulated as follows:
\begin{equation}
    P^{cls}_{s}(x)^{(i)} = 
    \frac{exp(g_{\theta_s}(x)^{(i)}_{cls} / \tau_s)}
    {\sum_{k}exp(g_{\theta_s}(x)^{(k)}_{cls} / \tau_s)}.
\end{equation}

Similarly, the teacher's output probability is denoted as $P^{cls}_{t}(x)$ with temperature $\tau_t$ and network $\theta_t$. During training, the given image is randomly augmented several times to construct a set $V$ of different distorted views. When local views are included by multi-crop augmentation, only global
views $\{x_1^g, x_2^g\}$ are passed through the teacher and all views are passed through the student. The vanilla DINO minimizes the cross-entropy loss between the output class token of the student and teacher:
\begin{equation}\label{loss_cls}
    \mathcal{L}_{cls} = \sum_{x\in {x_1^g,x_2^g}} \sum_{x' \in V, x' \neq x}
    -P^{cls}_t(x) log P^{cls}_s(x').
\end{equation}

\subsection{The Proposed DenseDINO}
\label{sec:ref}

Though DINO performs very competitively on image-level classification tasks, it performs poorly on pixel-level segmentation tasks as it only constrains the image-level cross-view consistency but neglects the spatial information. To remedy this, we propose to introduce point-level consistency on the basis of DINO to obtain a new framework called DenseDINO to force the model to perceive position-aware local semantic information.
Specifically, point-level supervision is implemented by minimizing the distillation loss between cross-view reference token pairs that share the same spatial position information in an image. The generation of reference tokens and the formulation of the loss function is as follows.

As is shown in Figure~\ref{fig:pipeline}, DenseDINO follows the overall framework of DINO except for the reference token. 
At the input stage, for each pair of views in set $V$, we randomly sample $M$ points in their overlapped area. 
Suppose that the absolute coordinate of a reference point is $(x,y)$ and the bounding box of the corresponding view is $(x_1, y_1, x_2, y_2)$. We then compute the relative coordinate of the reference point as $((x-x_1)/(x_2-x_1), (y-y_1)/(y_2-y_1))$. Following DeiT\cite{deit}, through reshaping the sequence of patch position embedding into a grid, we can encode the reference points into positional embeddings by bicubic interpolation according to their relative coordinates.

The obtained point position embeddings are regarded as input tokens, namely reference tokens, and fed into transformer blocks together with both the class token and patch tokens. Since reference points may appear anywhere in the image, we expect them to locate different objects in the image, and the model is forced to pay attention to multiple objects rather than the most salient one, which enriches the learned visual representation and improve the model's localization ability. 
At the output stage, similar to the class token, the output reference tokens are projected by a shared projection head and normalized with softmax function to obtain point-level predicted probability $P^{ref_j}_{s}(x)$ and $P^{ref_j}_{t}(x)$, where $ref_j$ refers to the $j$-th reference token.
And we formulate the point-level similarity loss of reference tokens, denoted as reference loss $\mathcal{L}_{ref}$, served as a complement to image-level similarity loss of class token $\mathcal{L}_{cls}$:
\begin{equation}
    \mathcal{L}_{ref} = 
    \sum_{x\in {x_1^g,x_2^g}} 
    \sum_{x' \in V, x' \neq x}
    \sum_{j=1}^{M}
    -P^{ref_j}_t(x) log P^{ref_j}_s(x').
\end{equation}

To this end, the model is trained by minimizing:
\begin{equation} \label{eq:final}
    \mathcal{L}_{total} = \alpha \times \mathcal{L}_{cls} + (1-\alpha) \times \mathcal{L}_{ref},
\end{equation}
where $\alpha$ is a hyper-parameters to balance the weights of two items, $\mathcal{L}_{cls}$ denotes the vanilla DINO loss in Eq.~(\ref{loss_cls}).

However, directly adding the reference tokens into the framework would bring the following two issues. On the one hand, at the early stage of training, the patch position embedding has not learned spatial relations yet. 
It is observed that introducing inaccurate reference tokens makes the training unstable. On the other hand, we expect to disentangle the reference tokens from other transformer modules and keep the inference process unchanged for easy transfer on downstream tasks.

\begin{figure}[t]
    \centering
    \includegraphics[width=0.6\linewidth]{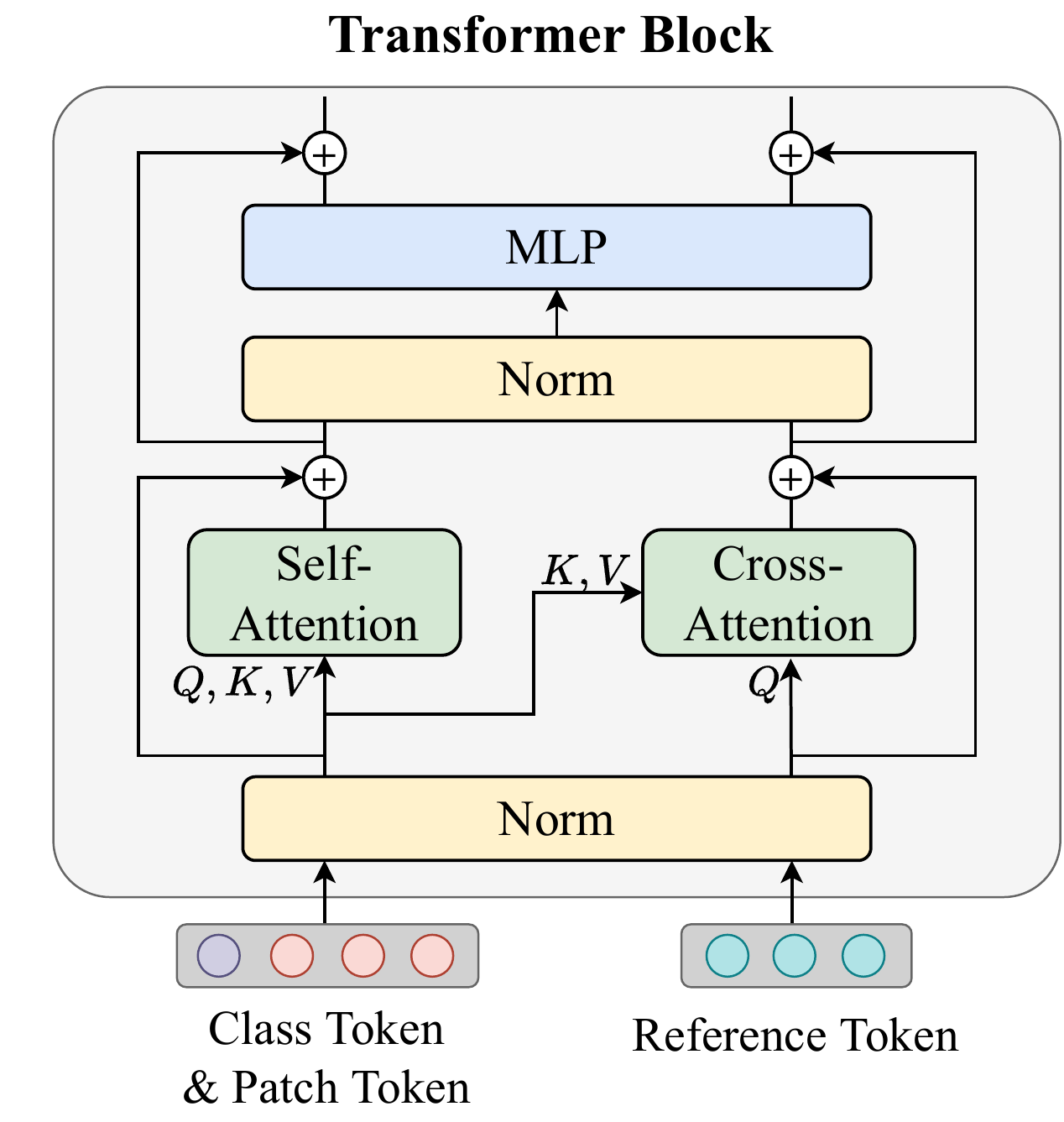}
    \caption{An overview of the modified transformer block in which reference token gathers information via a cross-attention module.} 
    \label{fig:mask}
\end{figure}

Therefore, in each transformer block, we replace the vanilla self-attention module with a modified masked-attention module. As shown in Figure \ref{fig:mask}, both the class token and patch tokens are processed with self-attention, while the reference tokens induct information from patch tokens via a cross-attention module. It is worth noting that the reference tokens are only used as queries and do not interact with each other, which guarantees the independency of each reference token. In the practical implementation, we simply transform self-attention into cross-attention without modifying the transformer block, but by building a mask that abandons corresponding columns of the attention matrix and passing it to each block as a parameter. In summary, DenseDINO introduces point-level consistency by introducing extra reference tokens and reformulates the attention module to make it more friendly to reference tokens at random positions.

\begin{figure*}[htbp]
    \centering
    \includegraphics[width=\linewidth]{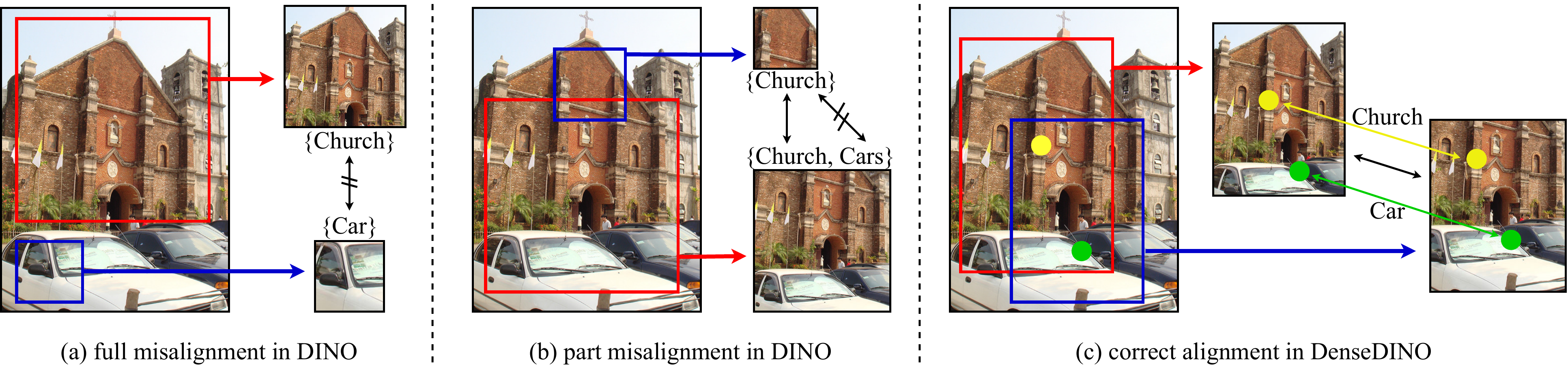}
    \caption{An example of two types of misalignment under image-level consistency in DINO and a case of multi-object alignment with point-level consistency in DenseDINO.} 
    \label{fig:multi-crop}
\end{figure*}

\subsection{Impact of Object Misalignment in Multi-Crop on Dense Prediction Tasks}
\label{sec:multi-crop}

Multi-crop\cite{swav} is a generic augmentation scheme applied to many self-supervised approaches. Multi-crop generates multiple views, including two global views of high resolution that cover large parts of the image and several local views of low resolution that cover small parts. As in Equation \ref{loss_cls}, the feature encoded from one global view is aligned with features from the other global view and all the local views. The mixture of multiple views provides more information included in the image and indeed brings significant performance gain for the classification task.

However, it is observed that multi-crop would hurt the segmentation performance significantly (see ablation study in Section \ref{sec:abl-multi-crop}). We speculate that this is because the usage of local view increases the probability of object misalignment, which indicates that the object information in different views is unequal.

To verify this, we divide object misalignment into two types: full misalignment and part misalignment.
When the global view and local view have no intersection, full misalignment may occur. As shown in Figure \ref{fig:multi-crop}(a), the pair views contain completely different objects.

And part misalignment, as shown in Figure \ref{fig:multi-crop}(b), refers to the case that there are multiple objects in the global view while the local view contains not all of them. 

The probability of an object being cropped in a view is proportional to two variables: the area of the object and the area of view. For the salient object, it is of high probability of being cropped in both global and local views because its area is large.  

Thus salient object easily maintains consistency between global and local views. 
However, for a small object, its probability of being cropped in the global view is high while being cropped in the local view is relatively low, leading to an increased probability of which small object only exists in the global view, {\it i.e.}, the part misalignment we mentioned above. 
Since image-level features of each view are aligned in DINO, to satisfy the consistency constraint, the model is forced to only encode the salient object that appears in both views and ignore those small objects that suffer from part misalignment.
Moreover, with the increase in the number of views, the salient object is more likely to dominate across views, further aggravating the misalignment. 
Therefore, multi-crop benefits classification since the salient object is more likely to be category-related and results in performance degradation on dense prediction tasks since information of all objects is required.

With the analysis above, we make the following modification in our DenseDINO framework.
Firstly, a lower bound for the overlapped area between each pair of views is set, which ensures that there exist objects included by both views and avoids full misalignment. 
Secondly, we replace the local views with global views. Alignment between global views makes it less likely to match multiple objects with a single one, which mitigates the part misalignment issue. 
Thirdly, reference tokens are applied to all pairs of views as shown in Figure \ref{fig:multi-crop} (c).
Compared to the class token which mainly captures the global semantic information, supervision provided with reference token are more accurate and sensitive to the spatial relationship. On the one hand, it will query those small objects when reference points are sampling on their positions. 
On the other hand, due to the position prior, the misalignment between point-level features never occurs. 
More analysis and experiments are provided in Section \ref{sec:abl-multi-crop}.

\section{Experiment}
\label{sec:exp}

\subsection{Experimental Setup}

\subsubsection{Implementation Details}
We choose ViT\cite{vit} as the backbone and ImageNet\cite{imagenet} as the training dataset. We train ViT-Small with 4 views and 4 reference points in each pair of views for 300 epochs for the performance comparison in main results, and ViT-Tiny with 6 views and 4 reference points for 100 epochs for the ablation study. The loss weight $\alpha$ is set as 0.5.
The projection head, centering, and sharpening settings of the reference token are exactly the same as the class token.
All models in the experiment are with patch size 16 and trained from scratch unless specified otherwise. Other training parameters are kept the same with the setting of DINO. 

\subsubsection{Datasets and Evaluation Metrics}
In this section, we evaluate our method on both classification and semantic segmentation tasks. To better illustrate the quality of the learned vision representation directly, we use evaluation protocols in which the model is frozen.
For classification, following \cite{knn}, we adopt a simple weighted k-NN classifier. During the evaluation, we first compute and store the features of the training images, match the feature of testing images to the $k$ nearest stored features, and vote for the label. The accuracy is reported by setting $k$ to 20. For semantic segmentation, we adopt linear probing protocol following \cite{leopart}. We train a 1$\times$1 convolutional layer on the frozen patch tokens on Pascal VOC 2012\cite{pascal} $train+aug$ split and report mIOU on the $valid$ split. The classification and semantic segmentation tasks are abbreviated as "Cls." and "Seg." respectively in the following.

\subsection{Main Results}
\label{sec:main}

\begin{table}
    \centering
    \begin{tabular}{crrr}
        \toprule
        Method  & Epochs & Cls. & Seg. \\
        \midrule
        \multicolumn{1}{l}{\textit{Image-level Approach:}}  & & \\
        Mocov2\cite{mocov2}     & 800   & 64.4  & 45.0  \\     
        SwAV\cite{swav}         & 800   & 66.3  & 49.0  \\
        DINO\cite{DINO}         & 800   & \textbf{74.4}  & 50.6  \\
        \midrule
        \multicolumn{1}{l}{\textit{Pixel/Patch-level Approach:}}  & & \\
        DenseCL\cite{densecl}   & -     & -     & 49.0  \\
        Leopart\cite{leopart}   & 800+50& 52.0  & \textbf{68.9}  \\
        \midrule
        \multicolumn{1}{l}{\textit{Our Reproduction Results:}}  & & \\
        DINO*       & 300       & \textbf{70.4}  & 56.1  \\
        Leopart*    & 300+50    & 61.0  & \textbf{64.6}  \\
        DenseDINO(Ours)     & 300       & 69.7  & 63.3  \\
        \bottomrule
    \end{tabular}
    \caption{Comparison (\%) with other methods. * are run by us with the official release codes and other results are referenced from prior work. All methods use ViT-Small of patch size 16 as the backbone. "Epochs" refers to the number of training epochs on ImageNet. Different from other methods, Leopart starts from DINO initialization and finetunes for additional 50 epochs.}
    \label{tab:main}
\end{table}

\begin{table}[t]
    \begin{tabular}{@{\hspace{0.1cm}}l@{\hspace{0.1cm}}|c@{\hspace{0.1cm}}c@{\hspace{0.1cm}}|r@{\hspace{0.25cm}}r}
        \toprule
        Method & Global Views & Local Views & Cls. & Seg. \\
        \midrule
        \multirow{4}{*}{DINO} & $2\times224$ & - & 53.3 & 46.0 \\
         & $2\times224$ & $4\times96$ & \textbf{58.8} & 39.7 \\
         & $2\times224+4\times96$ &  & 56.1 & 42.6 \\
         & $6\times224$ & - & 56.7 & \textbf{46.4} \\
        \midrule
        \multirow{4}{*}{DenseDINO} & $2\times224$ & - & 53.8 & 53.1 \\
         & $2\times224$ & $4\times96$ & \textbf{58.5} & 49.0 \\
         & $2\times224+4\times96$ & - & 57.5 & 51.4 \\
         & $6\times224$ & - & 55.7 & \textbf{57.4} \\
        \bottomrule
    \end{tabular}
    \caption{Ablation study (\%) on different settings of multi-crop. The global view is generated with a large crop scale (40\%$\sim$100\%) and the local view with a small scale (5\%$\sim$40\%). 
    "$2\times224$" indicates two views resized to resolution 224 after crop.}
    \label{tab:ablate-mc}
\end{table}

To illustrate the effectiveness of our proposed method, we compare the performance of our method with the existing self-supervised approaches on both classification and segmentation downstream tasks. 
For a fair comparison, we select DINO, the state-of-the-art method on the image-level prediction, and Leopart, the state-of-the-art method on the dense prediction, and reproduce their results by ourselves with the officially released codes. Due to the limited computational resources, we train DINO for 300 epochs with 4 views (the best model is trained for 800 epochs with 12 views) and finetune the model following Leopart for 50 epochs. And we use the exactly same setting of DINO but with our proposed reference token to train the model from scratch. As is shown in Table \ref{tab:main}, compared to the baseline DINO, we achieve competitive performance on ImageNet classification while surpassing DINO by 7\% on PascalVOC semantic segmentation. And compared to Leopart, without further finetuning, we obtain competitive performance on segmentation and outperform Leopart by 8\% on classification. It is worth noting that our method does not apply any direct supervision on patch tokens, which indicates that the performance improvement on segmentation comes from better visual representation rather than pretext-task prior.

\subsection{Ablation Study}
In this subsection, we conduct ablation studies on multi-crop, loss items, and the number of reference points.

\subsubsection{Analysis of Multi-crop}
\label{sec:abl-multi-crop}

In this section, we conduct an elaborate ablation study to understand multi-crop. There are two variables in the setting of multi-crop: crop scale (global view vs. local view) and view resolution (224 vs. 96). From the DINO results in Table \ref{tab:ablate-mc}, it is observed that with local views, the classification accuracy increases from 56.1\% to 58\%, while at the same time, the segmentation mIoU falls from 42.6\% to 39.7\%.
And with low resolution, the performance of the model degenerates on both tasks from 56.7\%, 46.4\% to 56.1\%, 42.6\%, respectively. To this end, we speculate that the local view brings the benefits of misalignment to classification at the cost of segmentation performance, using views with low resolution only hurts the generalization ability of learned features. Replacing local views of low resolution with global views of high resolution improves representation quality but suffers from bigger computational budgets. From the DenseDINO results, we observe that our DenseDINO could improve the segmentation performance significantly while keeping the classification accuracy competitive in all view settings, which attributes to the point-level consistency introduced by reference tokens.

Besides, it is worth noting that, similar to DINO, DenseDINO also performs poorly with standard multi-crop (see the sixth line of the table). With four additional local views, the segmentation performance of DenseDINO drops from 53.1\% mIoU to 49.0\% mIoU. We speculate that this is because reference tokens are sensitive to position and thus require accurate positional embedding. The usage of local views introduces extra noise via interpolating all patch position embedding and thus conflicts with localizing objects by reference tokens.

\begin{table}[t]
    \centering
    \begin{tabular}{cc|rr}
        \toprule
        $\mathcal{L}_{cls}$ & $\mathcal{L}_{ref}$  & Cls. & Seg. \\
        \midrule
         \checkmark &  & \textbf{56.7} & 46.4 \\
         & \checkmark & 42.4 & 55.9 \\
         \checkmark & \checkmark & 55.7 & \textbf{57.4} \\
        \bottomrule
    \end{tabular}
    \caption{Ablation study (\%) on vanilla DINO loss and reference loss.}
    \label{tab:ablation0}
\end{table}

\begin{figure}[t]
    \centering
    \includegraphics[width=\linewidth]{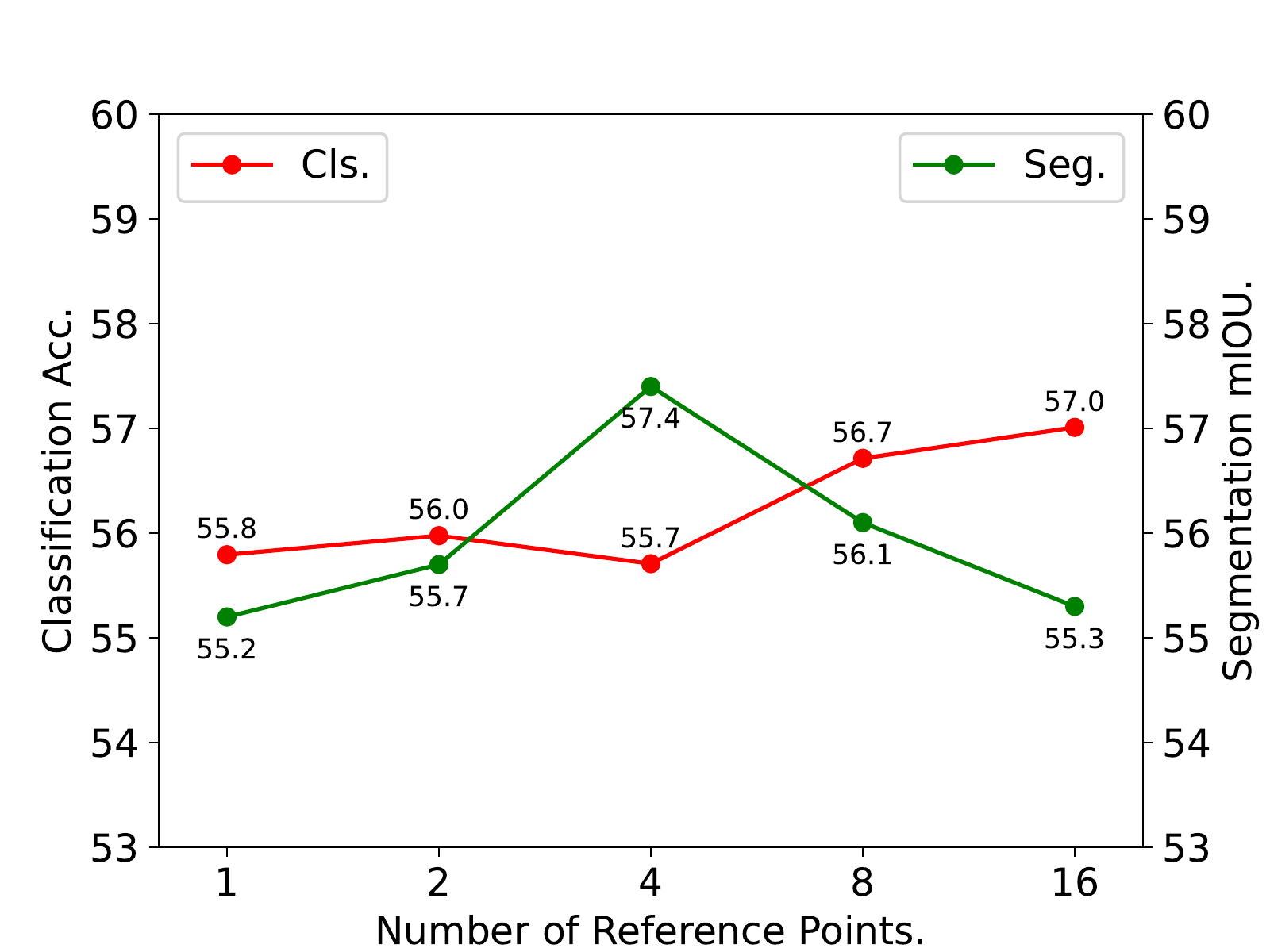}
    \caption{Ablation study (\%) on the number of reference points. The x-axis is in log-scale.} 
    \label{fig:num-point}
\end{figure}

\begin{figure*}[htbp]
    \centering
    \includegraphics[width=0.9\linewidth]{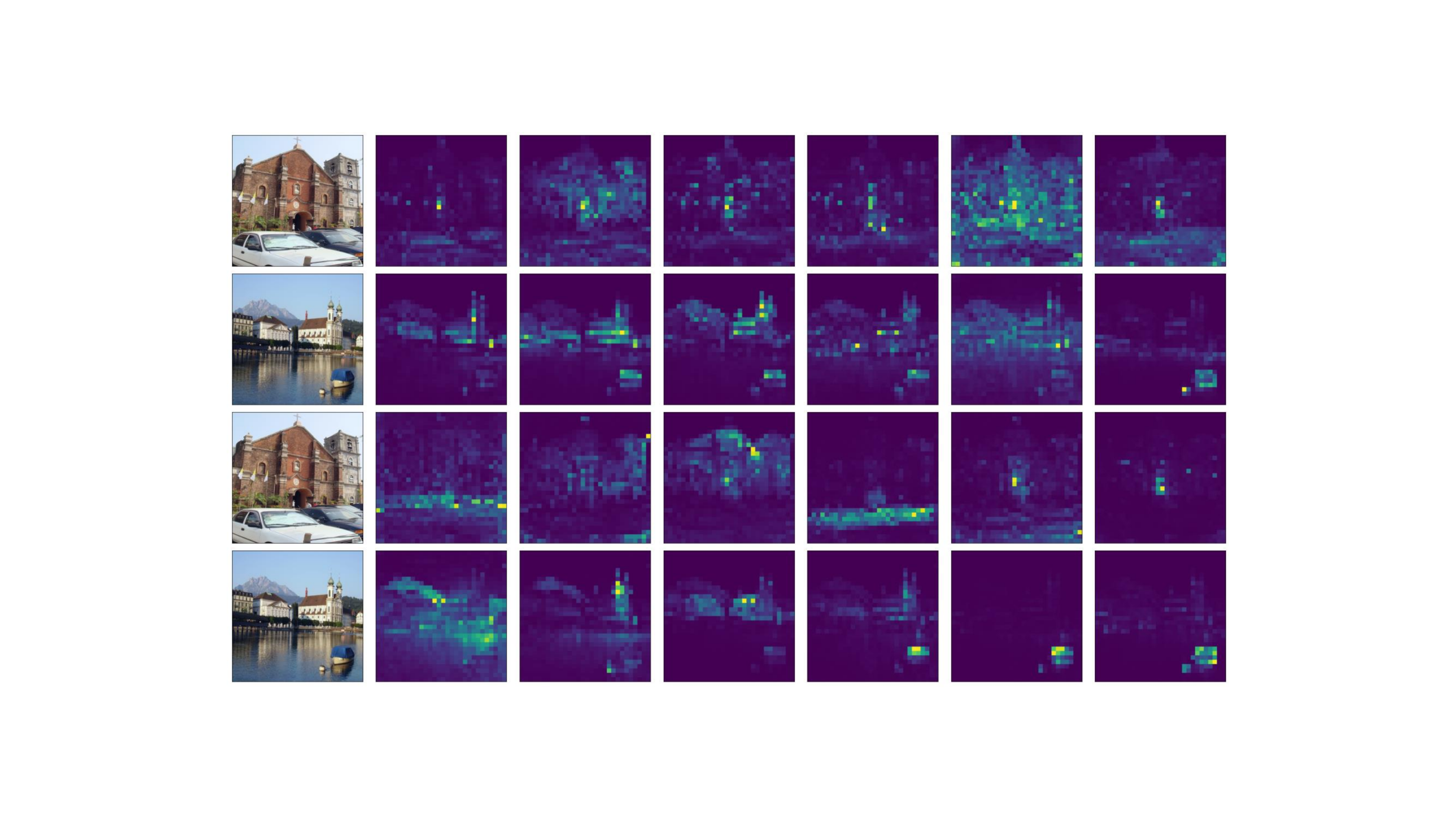}
    \caption{Visualization of multi-head attention of the class token in the last block of ViT-Small. We select images with multiple semantic objects (\textit{church/car} and \textit{church/boat}). The top two rows are the results of the baseline method (DINO) and the bottom two rows are the results of our method.}
    \label{fig:headatt}
\end{figure*}

\subsubsection{Ablation Study on Loss Items}

To evaluate the impacts of two similarity losses in Eq.~(\ref{eq:final}), we conduct ablation studies with different numbers of views. As shown in Table (\ref{tab:ablation0}), the vanilla DINO loss helps the model to capture global semantic information and thus benefits classification, the reference loss maximizes point-level consistency and improves segmentation performance. 

Furthermore, we notice that a simple hybrid training strategy of combining the two losses efficiently exploits the strengths of both approaches (-1\% on classification accuracy and +1.5\% on segmentation mIoU), which indicates that the reference tokens perform as the suitable partner for image-level approach.

\subsubsection{Effective Number of Reference Points}

We study the number of reference points $M$ sampled for each pair of views. Suppose there are $N_V$ views in the view set $V$, then $(N_V - 1) M$ reference points are sampled on each view. The number of views is 6. The results are shown in Figure \ref{fig:num-point}. It is observed that, as the number of reference points increases, the model performance on the classification task is getting better, which suggests that reference tokens complement the class token by better leveraging the information in the image. Moreover, we find that four reference points per view are sufficient for learning dense representations. 
Although more reference points bring more supervision signals, it may also cause the number of queries for the salient object to be much greater than that for small objects.
Since we average the losses of all reference tokens to form $\mathcal{L}_{ref}$, the dominance of the salient object is an obstacle to understanding complex scenes. Thus the performance of dense prediction tasks degenerates when a large number of reference tokens are used.

\subsubsection{Visualization of Attention Maps on Multi-Object Images}
As shown in Figure~\ref{fig:headatt}, we 
visualize the attention map of the class token in the last block of ViT-Small. We select two images with different objects (\textit{church/car} and \textit{church/boat}). Each column except the first one represents attention maps of a head in the multi-head attention. The top two rows are the results of the baseline method (DINO) and the bottom two rows are the results of our DenseDINO. It can be seen from the attention maps that almost all heads of DINO are related to the salient object (\textit{church}), which means that the baseline model mainly focuses on the salient object in a multi-object image. In contrast, our DenseDINO could capture more diverse semantic information, as both the salient object and the other semantic objects (\textit{car/boat}) are related in our attention maps. A potential reason for the attention results is that our method forces the model to capture the semantic features of different regions by the consistency of random cross-view reference point pairs, and the baseline model may take other parts except for the salient object in an image as insignificant parts without the point-level consistency constraints. This demonstrates that our DenseDINO is sensitive to different semantic parts in an image, which is required for the dense prediction tasks. 

\section{Conclusion}

In this work, we have proposed a dense self-supervised framework DenseDINO, which supervises the point-level cross-view consistency in a token-based way.
We have found that previous multi-crop augmentation causes an object misalignment problem between views and provided our analysis.
Extensive experiments demonstrate that DenseDINO improves the performance on dense prediction tasks by a large margin and remains comparable performance on the image-level prediction tasks.
In the future, we shell further optimize the selection and generation of reference tokens for better object localization and more accurate supervision.

\appendix



\clearpage

\section*{Acknowledgments}

This work is supported in part by National Key Research and Development Program of China under Grant 2020AAA0107400, National Natural Science Foundation of China under Grant U20A20222, National Science Foundation for Distinguished Young Scholars under Grant 62225605, the Ng Teng Fong Charitable Foundation in the form of ZJU-SUTD IDEA Grant, 188170-11102, CAAI-HUAWEI MindSpore Open Fund, as well as CCF-Zhipu AI Large Model Fund (CCF-Zhipu202302).

\bibliographystyle{named}
\bibliography{ijcai23}

\end{document}